\documentclass[11pt,a4paper]{article}
\usepackage[hyperref]{emnlp2020}
\usepackage{times}
\usepackage{latexsym}
\usepackage{url}
\usepackage{amsmath}
\usepackage{enumitem}
\setlist{nosep} 
\usepackage{CJKutf8}
\usepackage{makecell}

\usepackage{xcolor}
\usepackage{colortbl}

\aclfinalcopy 

\usepackage{graphicx}
\graphicspath{ {./images/} }

\title{Solving Historical Dictionary Codes with a Neural Language Model}

\author{Christopher Chu, Raphael Valenti, and Kevin Knight  \\
DiDi Labs \\
4640 Admiralty Way \\
Marina del Rey, CA 90292 \\
{\tt \{chrischu,kevinknight\}@didiglobal.com}}

\date{}

\begin{document}
\maketitle

\begin{abstract}
We solve difficult word-based substitution codes by constructing a decoding lattice and searching that lattice with a neural language model.  We apply our method to a set of enciphered letters exchanged between US Army General James Wilkinson and agents of the Spanish Crown in the late 1700s and early 1800s, obtained from the US Library of Congress.  We are able to decipher 75.1\% of the cipher-word tokens correctly.
\end{abstract}

\section{Introduction} 

Cryptography has been used since antiquity to encode important secrets.  There are many unsolved ciphers of historical interest, residing in national libraries, private archives, and recent corpora collection projects \cite{decode19,beata19}.  Solving classical ciphers with automatic methods is a needed step in analyzing these materials.

In this work, we are concerned with automatic algorithms for solving a historically-common type of {\em book code}, in which word tokens are systematically replaced with numerical codes. Encoding and decoding are done with reference to a dictionary possessed by both sender and recipient.  While this type of code is common, automatic decipherment algorithms do not yet exist.  The contributions of our work are:

\begin{itemize}
\item We develop a algorithm for solving dictionary-based substitution codes.  The algorithm uses a known-plaintext attack (exploiting small samples of decoded material), a neural language model, and beam search.
\item We apply our algorithm to decipher previously-unread messages exchanged between US Army General James Wilkinson and agents of the Spanish Crown in the late 1700s and early 1800s, obtaining 72.1\% decipherment word accuracy.
\end{itemize}

\section{Related Work}

\begin{figure}
\small
\begin{tabular}{l|l|l|}
& {\bf Table}-based key & {\bf Book}-based key \\ \hline
{\bf Cipher}  & Caesar cipher & Beale cipher \\
(character) & Simple substitution &  \\ 
 & Zodiac 408 &  \\
 & Copiale cipher &  \\ \hline
{\bf Code}  & Rossignols' & Mexico-Nauen code \\
(word) & \ \ Grand Chiffre & Scovell code \\ 
 & & Wilkinson code \\ 
 & & {\em \ \ (this work)} \\ \hline
\end{tabular}
\caption{Simplified typology of substitution-based cryptosystems, with some examples.  Ciphers involve character-level substitutions (e.g, {\em f} $\rightarrow$ {\em q}), while codes involve word-level substitutions (e.g., {\em forest} $\rightarrow$ {\em 5731}).}
\label{typology}
\end{figure}

Figure~\ref{typology} gives a simplified typology of classical, substitution-based cryptosystems.\footnote{Our typology is geared toward explaining our contribution in the context of related systems.  For a fuller picture of classical cryptology, the reader is directly to \newcite{kahn96} and \newcite{singh00}.  For example, we do not discuss here systems in which a substitution key evolves during the encoding process, such as the Vigen\`ere cipher or the German Enigma machine.}

{\bf Table-based Ciphers} involve character-based substitutions.  The substitution may take the form of a simple offset, as in the Caesar substitution system, e.g., ({\em a}~$\rightarrow$~{\em d}), ({\em b} $\rightarrow$ {\em e}), ({\em c} $\rightarrow$ {\em f}), etc.  The Caesar cipher can be easily solved by algorithm, since there are only 26 offsets to check.  The algorithm need only be able to recognize which of the 26 candidate plaintexts form good English.  Since 25 of the candidates will be gibberish, even the simplest language model will suffice.

A {\em simple substitution cipher} uses a substitution table built by randomly permuting the alphabet.  Since there are 26! $\approx$ 4~$\cdot$~10$^{26}$ possible tables, algorithmic decipherment is more difficult.  However, there are many successful algorithms, e.g., \cite{hart94,knight99,kondrak06,olson07,ravi08,penn10}.  Many of these systems search for substitution tables that result in candidate plaintexts that score well according to a character n-gram language model \cite{shannon51}, and they use search techniques like hill-climbing, expectation-maximization, beam search, and exact search.  The main practical challenge is to decipher short messages.  In a very long ciphertext, it is easy to ``spot the {\em q}'' because it is always followed by the same cipher character, which we can immediately guess stands for plaintext {\em u}, and so forth.  

More sophisticated ciphers use {\em homophonic} substitution, in which plaintext characters are replaced non-deterministically.  By applying high nondeterminism to frequent characters, the cryptographer can flatten out ciphertext character frequencies. Homophonic ciphers occur frequently in historical collections.  The Copiale cipher \cite{copiale} is a well-known example from a German secret society in the 1700s.  These ciphers can also be attacked successfully by algorithm.  For example, the homophonic Zodiac 408 cipher can be solved with EM with restarts \cite{berg13}, Bayesian sampling \cite{ravi11}, or beam search \cite{malte14} (all with n-gram character language models).  \newcite{sarkar18} employ a more powerful character-based neural language model to break short ciphers more accurately.  In the present work, we use a word-based neural language model.

{\bf Book-based ciphers} increase homophony, and also avoid physical substitution tables that can be stolen or prove incriminating.  In a book-based cipher, sender and recipient verbally agree up front on an innocuous-looking shared document (the ``book''), such as the US~Declaration of Independence, or a specific edition of the novel {\em Moby Dick}.  When enciphering a plaintext letter token like {\em f}, the sender selects a random letter {\em f} from the shared document---if it is the 712th character in the document, the plaintext {\em f} might be enciphered as {\em 712}.  The next plaintext {\em f} might be enciphered differently.

 \newcite{malte14} solve one of the most well-known book ciphers, part two of the Beale Cipher \cite{king93}.  Surprisingly, they treat the cipher as a regular homophonic cipher, using the same beam-search algorithm as for the table-based Zodiac 408 cipher, together with an 8-gram character language model.  One might imagine exploiting the fact that the book is itself written in English, so that if ciphertext unit {\em 712} is known to be {\em f}, then ciphertext unit {\em 713} is probably not {\em h}, as {\em fh} is unlikely to appear in the book.  \newcite{malte14}'s simple, effective algorithm ignores such constraints.  Other methods have been proposed for attacking book ciphers, such as {\em crib dragging} \newcite{churchhouse}.
 
{\bf Codes}, in contrast to ciphers, make substitutions at the whole-word level.\footnote{Commonly, a single historical system will mix letter substitutions and word substitutions. Such a system is called a {\em nomenclator}.}  A large proportion of the encrypted material in \newcite{decode19} consists of {\bf table-based codes}. A famous example is Antoine Rossignol's {\em Grand Chiffre}, used during the reign of Louis~XIV.  The sender and receiver each own copies of huge specially-prepared tables that map words onto numbers (e.g., {\em guerre}~$\rightarrow$~{\em 825}).  If the enciphering tables are kept secret, this type of code is very hard to break.  One might guess that the most frequent cipher token stands for the word {\em the}, but it quickly becomes challenging to decide which number means {\em practice} and which means {\em paragon}.  Even so, \newcite{dou12} take on the task of automatically deciphering newswire encrypted with an arbitrary word-based substitution code, employing a slice-sampling Bayesian technique.  Given a huge ciphertext of $\sim$50,000 words, they can decipher $\sim$50\% of those tokens correctly.  From one billion ciphertext tokens, they recover over 90\% of the word tokens.  However, this method is clearly inapplicable in the world of short-cipher correspondence.

In the present work, we consider {\bf book-based codes}. Instead of using specially-prepared tables, the sender and receiver verbally agree to use an already-existing book as a key.  Because it may be difficult to find a word like {\em paragon} in a novel like {\em Moby Dick}, the sender and receiver often agree on a shared pocket dictionary, which has nearly all the words.  If {\em paragon} were the 10,439th word in the dictionary, the sender might encode it as {\em 10439}.  

Such codes have been popular throughout history, employed for example by George Scovell during the Napoleonic Wars \cite{urban02}, and by John Jay during the US~Revolutionary War \cite{johnjay}.  They were used as late as World War~II, when German diplomats employed the {\em Langenscheidt's Spanish-German Pocket Dictionary} as a key to communicate between the cities of Chapultepec, Mexico and Nauen, Germany \cite{nauen}.  In that case, the US~Coast Guard intercepted messages and was able to make a bit of headway in deciphering them, but the real breakthrough came only when they obtained the applicable dictionary (key).

Unfortunately, there appear to be no automatic algorithms for solving book-based codes without the key.\footnote{\newcite{kahn96} suggests that national security services have long ago digitized all published books and applied brute-force to find the book that renders a given code into natural plaintext.}  According to \newcite{schmeh20}:

\begin{quotation}
\small
\noindent
``So far, there are no computer programs for solving codes and nomenclators available.  This may change, but in the time being, solving a code or nomenclator message is mainly a matter of human intelligence, not computer intelligence.''
\end{quotation}

In this paper, we develop an algorithm for automatically attacking book-based codes, and we apply it to a corpus of historically-important codes from the late 1700s.

\section{Wilkinson Letters}

\begin{table}
\begin{small}
\begin{center}
\begin{tabular}{|l|r|r|} \hline
& Word tokens & Word types \\ \hline
Deciphered letters &  approx.~800   & 326 \\ \hline
Evaluation set & 483 & 226 \\ \hline
Test set & 341 & 192 \\ \hline
\end{tabular}
\end{center}
\end{small}
\caption{Summary of transcribed data.  Deciphered and evaluation sets have gold-standard decipherments; the evaluation set is held out.  The test set has no gold-standard decipherment.}
\label{data-table}
\end{table}


\begin{figure*}
\begin{center}
\includegraphics[scale=0.4]{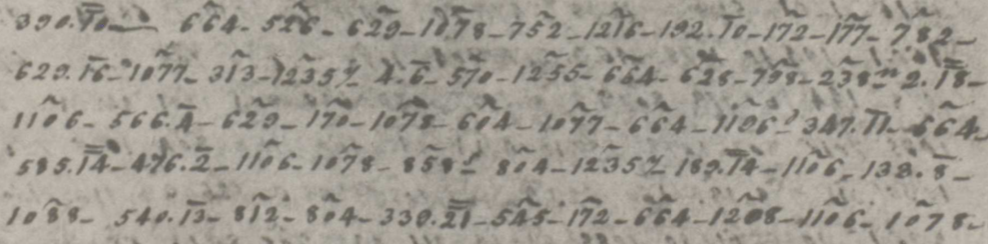}
\end{center}
\begin{scriptsize}
\begin{verbatim}
390.[10]= . [664]^ [526]^ [629]^ [1078]^ [752]^ [1216]^ 192.[10]- [172]^ [177]^ [782]^ 
629.[16]- [1077]^ [313]^ [1235]^ +y 4.[6]- [570]^ [1255]^ [664]^ [628]^ [798]^ [238]^ +n 2.[18]= 
[1106]^ 566.[4]- [629]^ [170]^ [1078]^ [604]^ [1077]^ [664]^ [1106]^ 347.[11]- [664]^ 
585.[14]= 476.[2]- [1106]^ [1078]^ [858]^ [804]^ [1235]^ +y 189.[14]= [1106]^ 133.[8]- 
[1088]^ 540.[13]- [812]^ [804]^ 339.[21]= [545]^ [172]^ [664]^ [1208]^ [1106]^ [1078]^ 

                                                     (a)
\end{verbatim}
\end{scriptsize}

\begin{center}
\includegraphics[scale=0.39]{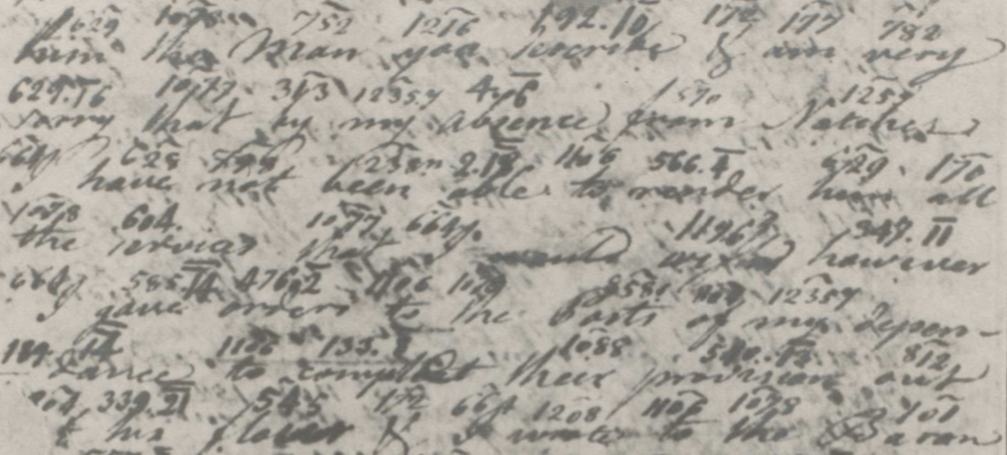}
\end{center}

\begin{small}
\begin{verbatim}
... him the man you described and am very 
sorry that by my absence from Natches
I have not been able to render him all 
the service that I wished however 
I gave orders to the ports of my dependence
to complete their provisions over 
of his ? and I wrote to the ...
\end{verbatim}
\end{small}
\begin{scriptsize}
\begin{verbatim}
                                                     (b)
\end{verbatim}
\end{scriptsize}

\caption{Sample encoded letters from US General James Wilkinson, with transcriptions:  (a) portion of a ciphertext letter, (b) a recovered intermediate version of this letter with both ciphertext and plaintext.}
\label{undeciphered}
\end{figure*}

Our cipher corpus consists of letters to and from US General James Wilkinson, who first served as a young officer in the US~Revolutionary War.  He subsequently served as Senior Officer of the US~Army (appointed by George Washington) and first Governor of the Louisiana Territory (appointed by Thomas Jefferson).  Wilkinson also figured in the Aaron Burr conspiracy \cite{burr}.  

Long after his death, letters in a Cuban archive revealed the famous Wilkinson to be an agent of the Spanish Crown during virtually his entire service, and his reputation collapsed \cite{linklater09}.


Table~\ref{data-table} summarizes our Wilkinson correspondence data.\footnote{All data is included with our released code (https://github.com/c2huc2hu/wilkinson/).}  We transcribe scans of manuscripts in the US Library of Congress.  We have 73pp of undeciphered text (Figure~\ref{undeciphered}a) and 28pp of deciphered text (Figure~\ref{undeciphered}b), with some overlap in content. Deciphered correspondence, with plaintext above ciphertext, likely resulted from manual encryption/decryption carried out at the time.

\section{Encryption Method}

As is frequent in book codes, there are two types of substitutions.  Some plaintext words are enciphered using a large shared table that maps words onto numbers ({\bf table-based code}).  Other words are mapped with a shared dictionary ({\bf book-based code}).  Despite serious efforts, we have not been able to obtain the dictionary used in these ciphers. 

In our transcription, we mark entries from the table portion with a caret over a single number, e.g., [123]\^{}. Before [160]\^{}, the table seems to contain a list of people or place names; between [160]\^{} (``a'') and [1218]\^{} (``your''), a list of common words in alphabetic order; and finally more common words. The last block was likely added after the initial table was constructed, suggesting that the table was used to avoid having to look up common words in the dictionary.

The ciphertext for the dictionary code has two numbers that mark a word's dictionary page and row number, respectively, plus one or two bars over the second number indicating the page column. For example, 123.[4]= refers to the fourth row of the second column of the 123rd page in the dictionary. From the distribution of cipher tokens, the dictionary is about 780 pages long with 29 rows per column, totaling about 45,000 words.

The cipher usually does not contain regular inflected forms of words, though inflections are sometimes marked with a superscript (e.g., $\hspace{0mm}^{+ing}$). 
Numbers and some words are left in plaintext. Long horizontal lines mark the ends of sentences, but other punctuation is not marked. 



\begin{figure}
\begin{small}
\begin{tabular}{|r|r|} \hline
\multicolumn{2}{|c|}{\bf Table wordbank} \\ \hline
Cipher & Plain  \\ \hline
[13]\^ & \tiny{Wilkinson} \\ \hline
[33]\^ & \tiny{Kentucky} \\ \hline
[92]\^ & \tiny{Philadelphia} \\ \hline \hline
[160]\^ & a \\ \hline
[172]\^ & and \\ \hline
[229]\^ & be \\ \hline
[231]\^ & bear \\ \hline
[313]\^ & by \\ \hline
... & ... \\ \hline
[1218]\^ & your \\ \hline \hline
[1235]\^ & me \\ \hline
[1249]\^ & policy \\ \hline
\end{tabular}
\parbox{.49\linewidth}{
\begin{tabular}{|r|r|r|} \hline
\multicolumn{3}{|c|}{\bf Dictionary wordbank} \\ \hline
Cipher & Plain & Index   \\ \hline
7.[24]-  & acquisition &  24 \\ \hline
15.[21]- & after & 485 \\ \hline
29.[29]- & answer & 1305 \\ \hline
44.[28]- & attache & 2174 \\ \hline
47.[21]- & attachment & 2341 \\ \hline 
59.[19]- & bearer & 3035 \\ \hline
65.[17]= & better & 3410 \\ \hline
75.[29]- & bosom & 3973 \\ \hline
103.[40]= & chamber & 5637 \\ \hline
113.[4]- & cipher & 6152 \\ \hline
114.[20]- & civil & 6226 \\ \hline
... & ... & ... \\ \hline
\end{tabular}
}
\end{small} 
\caption{Table wordbank (left) and dictionary wordbank (right) extracted from five deciphered letters.  The table contains proper nouns, an alphabetic word list (a--your) and other common words. Dictionary codes are of the form ``page.[row].col'', where ``-'' indicates column 1 and ``='' column 2.  The Index guesses that ``answer'' is the 1305\textsuperscript{th} word in the dictionary.} 
\label{wordbank1}
\end{figure}

\section{Automatic Decryption Method}

As the corpus includes a handful of deciphered pages, we employ a known-plaintext attack \cite{kahn96}.  

We first extract a small {\em wordbank} of known mappings, shown in Figure~\ref{wordbank1}. Next, we apply the wordbank to our held-out evaluation ciphertext.  We find that 40.8\% of word tokens can be deciphered, mainly common words.  After this step, we render a ciphertext as: 

\begin{small}
\begin{verbatim}
I ? he the man you ? and ? very ? that 
by ? ? from ? I have not ? ? to ? he ?
the service that I [...]
\end{verbatim}
\end{small}

\begin{figure*}[tb]
\centering
\includegraphics[width=6in]{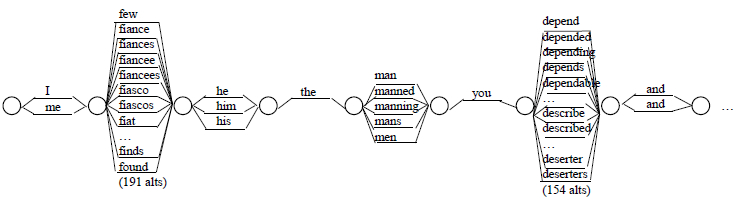}
\caption{Turning an encrypted letter into a lattice of possible decipherments.  Segments with few alternatives come from wordbank substitutions (and their automatically-produced morphological variants), while other segments come from interpolation-based guesses.  Each link has an associated probability (not shown here). On average, we supply 692 alphabetically-close alternatives per segment, but supply fewer than ten for most.}
\label{lattice}
\end{figure*}

\noindent
This is not yet a useful result.  However, the wordbank also helps us to recover the rest of the plaintext. Since both the table and dictionary are in alphabetical order, we use the wordbank to constrain the decipherment of unknown words.  

For example, given cipher word [163]\^{}, we know from Figure~\ref{wordbank1} that its plaintext must lie somewhere between the two anchor-words [160]\^{} (which stands for ``a'') and [172]\^{} (which stands for ``and'').  Moreover, it is likely to be closer to ``a'' than ``and''.  Repeating this for every cipher word in an undeciphered document, we construct a word lattice of all possible decipherments, shown in Figure~\ref{lattice}.  Our goal is then to search for the most fluent path through this lattice. Following are the details of our method:

{\bf Anchors.} To propose candidate words between two anchors,
we use a modern lemma-based dictionary with 20,770 entries.\footnote{www.manythings.org/vocabulary/lists/l (core ESL)}  In this dictionary, for example, there are 1573 words between ``attachment'' and ``bearer''.

\begin{figure}[t]
\begin{small}
\begin{center}
\includegraphics[scale=1.5]{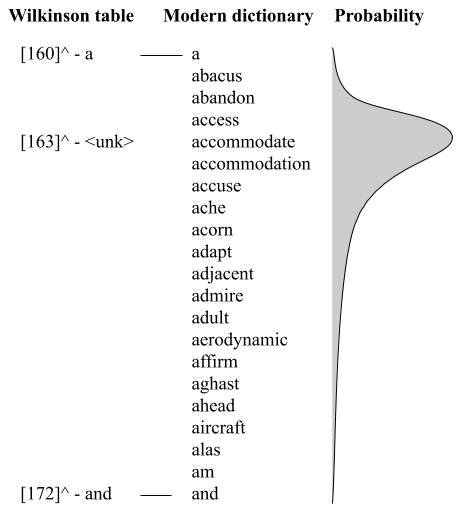}
\end{center}
\end{small}
\caption{Interpolating a ciphertext word not present in the wordbank. When deciphering [163]\^{}, we list candidate decipherments using a modern dictionary.  We assign probabilities to candidates based on interpolation between anchor words, in this case ``a'' and ``and''. }
\label{interpolate}
\end{figure}

{\bf Probabilities.} We assign a probability to each candidate based on its distance from the ideal candidate.  For example, in the table code, [163]\^{} is 30\% of the way from [160]\^{} to [172]\^{} (Figure~\ref{interpolate}), so our ideal candidate decipherment of [172]\^{} will be 30\% of the way between ``a'' and ``and'' in our modern dictionary.  To apply this method to the dictionary code, we convert each cipher word's page/column/row to a single number $n$ (the ``Index'' in Figure~\ref{wordbank1}), which estimates that the cipher word corresponds to the $n$th word in the shared dictionary.

We use a beta distribution for assigning probabilities to candidate words, because the domain is bounded.  We parameterize the distribution $B'(x;m,\beta)$ with mode $m$ and sharpness parameter $\beta$=5.  This is related to the standard parameterization, $B(x; \alpha, \beta)$, by:
\begin{equation*}
B'(x; m, \beta) = B\left(x;\alpha=\frac{m \beta - 2m + 1}{1 - m},\beta \right)
\end{equation*}

The sample space (0 to 1) is divided equally between the $M$ words in the modern dictionary, so the $i^{th}$ word gets probability: 

\begin{equation*}
\int^\frac{i}{M}_\frac{i-1}{M} B'(x; m, \beta) dx
\end{equation*}

In our example, because [163]\^{} is 30\% of the way from [160]\^{} to [172]\^{}, we have $m=0.3$. There are $M=650$ words in the modern dictionary between these two anchors, so the $i=105^{th}$ word (``access''), gets probability $0.00231$. 

{\bf Inflections.} We expand our lattice to include inflected forms of words (e.g., ``find'' $\rightarrow$ ``find'', ``found'', ``finding'').  
We generate inflections with the {\em Pattern} library \cite{DeSmedt:2012:PP:2188385.2343710}. Some words are generated more than once, e.g., ``found'' is both a base verb and the past tense of ``find''. {\em Pattern} inflects some uncommon words incorrectly, but such inflections are heavily penalized in the best-path step. 
Inflections divide the probability of the original word equally.

{\bf Table edge cases.} We replace unknown entries before the first anchor in the table with an arbitrary proper noun (``America''), and words outside the alphabetic section of the table with equal probability over a smaller vocabulary containing the 1000 most common words.\footnote{\scriptsize simple.wiktionary.org/wiki/Wiktionary:BNC\_spoken\_freq\_01HW}

{\bf Scoring lattice paths.}  After we have constructed the lattice, we automatically search for the best path.  The best path should be fluent English (i.e., assigned high probability by a language model), and also be likely according to our wordbank (i.e., contain high-probability lattice transitions).  

To score fluency, we use the neural GPT2 word-based language model \cite{radford2019language}, pre-trained on $\sim$40GB of English text.  We use the HuggingFace implementation\footnote{huggingface.co/transformers/pretrained\_models.html} with 12 layers, 768 hidden units, 12 heads, and 117M parameters.  

Neural language models have significantly lower perplexity than letter- or word-based n-gram language models.  For example, \newcite{raphael} benchmark WikiText-103 results for a Kneser-Ney smoothed 5-gram word model (test perplexity = 152.7) versus a quasi-recurrent neural network model (test perplexity = 32.8).   This gives neural language models a much stronger ability to distinguish good English from bad.

{\bf Beam search.}  We search for the best lattice path using our own beam search implementation.  To score a lattice transition with GPT, we must first tokenize its word into GPT's subword vocabulary. Since alphabetically-similar words often start with the same subword, we create a subword trie at each lattice position; when a trie extension falls off the beam, we can efficiently abandon many lattice transitions at once.



\section{Evaluation}

To evaluate, we hold out one deciphered document from wordbanking. Some plaintext in that document is unreadable or damaged, so decryptions are added when known from the wordbank or obvious from context.




Table~\ref{results} gives our results on per-word-token decipherment accuracy.  Our method is able to recover 73.8\% of the word tokens, substantially more than using the wordbank alone (40.8\%).  We also outperform a unigram baseline that selects lattice paths consisting of the most popular words to decipher non-wordbank cipher tokens (46.9\%).

The maximum we could get from further improvements to path-extraction is 91.3\%, as 8.7\% of correct answers are outside the lattice.  This is due to unreadable plaintext, limitation of our modern dictionary, use of proper names [1]\^{} to [159]\^{}, transcription errors, etc.

Table~\ref{beam} details the effect of beam size on decipherment accuracy, runtime, and path score (combining GPT log probability with lattice scores).  Increasing beam size leads us to extract paths with better scores, which correlate experimentally with higher task accuracy. 
\begin{table}
\begin{small}
\begin{center}
\begin{tabular}{|l|r|} \hline
\bf Method  & Token accuracy \\ \hline
Apply wordbank &  40.8  \\ \hline
Unigram baseline & 46.9  \\ \Xhline{2\arrayrulewidth}
Our method, beam size = 1 & 68.3  \\ \hline
Our method, beam size = 4 & 73.0  \\ \hline
\ \ + domain-tuning & 73.2  \\ \hline
\ \ \ \ + self-learning & {\bf 75.1} \\ \hline
Our method, beam size = 16 & 73.8  \\ \hline
\ \ + domain-tuning + self-learning & 75.1  \\ \Xhline{2\arrayrulewidth}
Oracle & 91.3 \\ \hline
\end{tabular}
\end{center}
\end{small}
\caption{Token decipherment accuracy.  The Unigram baseline selects the most frequent word at each point in the lattice.  Our method uses beam search to extract the path with highest GPT and lattices scores.  Oracle selects the path through the lattice that most closely matches the gold decipherment.} 
\label{results}
\end{table}

\begin{table}
\begin{small}
\begin{center}
\begin{tabular}{|r|r|r|r|} \hline
Beam size & Runtime & Internal Best & Task \\
&  &  Path Score & Accuracy \\ \hline
1  & 1m 38s & -3714.9 & 68.3 \\ \hline
4  & 6m 19s & -3569.6 & 73.0 \\ \hline
16  & 25m 31s & -3544.9 & 73.8 \\ \hline
\end{tabular}
\end{center}
\end{small}
\caption{Effect of beam size on decipherment. A larger beam lets us extract paths with better internal scores, which correlate with better task accuracy.} 
\label{beam}
\end{table}

Figure~\ref{solution} shows a portion of our solution versus the gold standard.  

For tokens where our system output does not match the original plaintext, we asked an outside annotator to indicate whether our model captures the same meaning.  For example, when our system outputs ``I am much sorry'' instead of ``I am very sorry,'' the annotator marks all four words as {\em same-meaning}. Under this looser criterion, accuracy rises from 73.0\% to 80.1\%.

We also decipher a Wilkinson ciphertext letter for which we have no plaintext. Transcription is less accurate, as we cannot confirm it using decipherment. The letter also includes phrases in plaintext French, which we translate to English before adding them to the lattice.  Despite these challenges, the model still outputs relatively fluent text, including, for example: ``\ldots as may tend most powerfully and most directly to dissolve the whole America of the first states from the east and to cease the intercourse of the west.''  This passage is consistent with Wilkinson's plan to seek independence for parts of America.

\section{Additional Methods}

We experiment with three additional decipherment methods.

\begin{table}
\begin{center}
\begin{small}

\begin{tabular}{|r|r|} \hline
{\em a} & Accuracy  \\ \hline
0.2 & 70.5 \\ \hline
0.5 & 71.6 \\ \hline
1 & {\bf 73.0} \\ \hline
2 & 72.2 \\ \hline
\end{tabular}
\parbox{.49\linewidth}{
\begin{tabular}{|r|r|} \hline
$\beta$ & Accuracy   \\ \hline
1  &  67.8 \\ \hline
3  &  71.6 \\ \hline
5  &  {\bf 73.0} \\ \hline
10  &  70.7 \\ \hline
\end{tabular}}
\end{small} 
\end{center}
\caption{Effects on decipherment accuracy of $a$ (weight applied to lattice scores vs.~GPT scores for each path) and $\beta$ (sharpness parameter for candidate-word probabilities).}
\label{varyingparams}
\end{table}

{\bf Weighted scoring}.  When scoring paths, we sum log probabilities from GPT and the lattice transitions, with the two sources equally weighted.  This turns out to be optimal.  Table~\ref{varyingparams} gives results when we multiply the lattice-transition score by $a$.  Halving the weight of the lattice scores degrades accuracy from 73.0 to 71.6 (-1.4 for beam=4), while doubling it degrades from 73.0 to 72.2 (-0.8 for beam=4).  Table~\ref{varyingparams} also shows the impact of the sharpness parameter $\beta$ on accuracy.

{\bf Domain-tuned language model.} We collect letters written by  Wilkinson\footnote{founders.archives.gov} totalling 80,000 word tokens, and fine-tune the GPT language model for one epoch on this data. The domain-tuned GPT increases decipherment accuracy from 73.0 to 73.2 (+0.2 for beam=4).  Fine tuning for more than one epoch degrades decipherment accuracy.  
We found experiments with  COFEA\footnote{https://lcl.byu.edu/projects/cofea/} (American English sources written between 1765 and 1799) to be fruitless. We fine-tune a language model on a COFEA subset consisting of 3.8 million word tokens for one epoch, but this degrades accuracy from 73.0\% to 65.6\%.

{\bf Iterative self-learning.} We apply iterative self-learning to improve our decipherment.  After extracting the best path using beam search, we take the words with the smallest increases in perplexity on the lattice and language models, and we add them to the wordbank.  The new wordbank provides tighter anchor points.  We then construct a new lattice (using the expanded wordbank), search it, and repeat. This further improves decoding accuracy to 75.1 (+1.9 for beam=4).

\begin{table}
\begin{small}
\begin{tabular}{|r|r|r|r|} \hline
Parallel & Wordbank & Coverage  & Decipherment \\ 
plain-cipher & size & of test & accuracy \\ 
tokens & & tokens & \\ \hline
500 & 227 & 53.8 & 66.0 \\ \hline
800 & 351 & 58.3 & 69.6 \\ \hline
2000 & 691 & 68.7 & 73.5 \\ \hline
20000 & 2904 & 76.6 & 79.6 \\ \hline
\end{tabular}

\end{small} 
\caption{Experiments with synthetic data.  By enciphering material from Project Gutenberg, we produce arbitrary-sized wordbanks from arbitrary amounts of parallel plaintext-ciphertext.  We then test how well those wordbanks support decipherment of new material.
}
\label{dataefficiency}
\end{table}

\section{Synthetic Data Experiments}

We next experiment with synthetic data to test the data efficiency of our method. To create arbitrary amounts of parallel plaintext-ciphertext data, we encipher a book from Project Gutenberg,\footnote{{\em Wisdom of Father Brown}, https://www.nltk.org/} using a different machine readable dictionary.\footnote{https://packages.ubuntu.com/xenial/wamerican}  We build wordbanks from parallel documents and use them to decipher a separately-enciphered book by the same author.\footnote{{\em The Man Who Was Thursday}, https://www.nltk.org/} The results are shown in Table~\ref{dataefficiency}.


\begin{figure*}
\small
\begin{center}
\begin{tabular}{ |l|l|l|l|l| }
\hline
Ciphertext & Gold & Wordbank only  & Unigram Baseline & Our method  \\
 \hline
[229]\^ {\tiny +ing} & being & be & be & {\cellcolor{lightgray} \bf being} \\ \hline
[186]\^ & at & {\cellcolor{lightgray} \bf at} & {\cellcolor{lightgray} \bf at} & {\cellcolor{lightgray} \bf at} \\ \hline
[1049]\^ & such & {\cellcolor{lightgray} \bf such} & {\cellcolor{lightgray} \bf such} & {\cellcolor{lightgray} \bf such} \\ \hline
[160]\^ & a & {\cellcolor{lightgray} \bf a} & {\cellcolor{lightgray} \bf a} & {\cellcolor{lightgray} \bf a} \\ \hline
212.[20]= & distance &  & distant & {\cellcolor{lightgray} \bf distance} \\ \hline
[570]\^ & from & {\cellcolor{lightgray} \bf from} & {\cellcolor{lightgray} \bf from} & {\cellcolor{lightgray} \bf from} \\ \hline
[90]\^ & North Carolina & America & America & America \\ \hline
[172]\^ & and & {\cellcolor{lightgray} \bf and} & {\cellcolor{lightgray} \bf and} & {\cellcolor{lightgray} \bf and} \\ \hline
286.[14]= & for & {\cellcolor{lightgray} \bf for} & {\cellcolor{lightgray} \bf for} & {\cellcolor{lightgray} \bf for} \\ \hline
[509]\^ & fear &  & father & {\cellcolor{lightgray} \bf fear} \\ \hline
[804]\^ & of & {\cellcolor{lightgray} \bf of} & {\cellcolor{lightgray} \bf of} & {\cellcolor{lightgray} \bf of} \\ \hline
446.[1]- {\tiny +ing} & missing &  & mistress & {\cellcolor{lightgray} \bf missing} \\ \hline
[1218]\^ & your & {\cellcolor{lightgray} \bf your} & {\cellcolor{lightgray} \bf your} & {\cellcolor{lightgray} \bf your} \\ \hline
294.[20]= & garrison &  & from & friend \\ \hline
[1084]\^ & there &  & therefore & therefore \\ \hline
286.[14]= & for & {\cellcolor{lightgray} \bf for} & {\cellcolor{lightgray} \bf for} & {\cellcolor{lightgray} \bf for} \\ \hline
[1078]\^ & the & {\cellcolor{lightgray} \bf the} & {\cellcolor{lightgray} \bf the} & {\cellcolor{lightgray} \bf the} \\ \hline
153.[5]- & conveyance &  & could & convenience \\ \hline
[804]\^ & of & {\cellcolor{lightgray} \bf of} & {\cellcolor{lightgray} \bf of} & {\cellcolor{lightgray} \bf of} \\ \hline
678.[6]= & this & {\cellcolor{lightgray} \bf this} & {\cellcolor{lightgray} \bf this} & {\cellcolor{lightgray} \bf this} \\ \hline
[664]\^ & I &  & {\cellcolor{lightgray} \bf I} & {\cellcolor{lightgray} \bf I} \\ \hline
[177]\^ & am &  & any & {\cellcolor{lightgray} \bf am} \\ \hline
467.[24]- {\tiny +d} & obliged &  & of & {\cellcolor{lightgray} \bf obliged} \\ \hline
[1106]\^ & to & {\cellcolor{lightgray} \bf to} & {\cellcolor{lightgray} \bf to} & {\cellcolor{lightgray} \bf to} \\ \hline
[1206]\^ & write & {\cellcolor{lightgray} \bf write} & written & {\cellcolor{lightgray} \bf write} \\ \hline
[1106]\^ & to & {\cellcolor{lightgray} \bf to} & {\cellcolor{lightgray} \bf to} & {\cellcolor{lightgray} \bf to} \\ \hline
[1216]\^ & you & {\cellcolor{lightgray} \bf you} & {\cellcolor{lightgray} \bf you} & {\cellcolor{lightgray} \bf you} \\ \hline
[807]\^ & in & on & {\cellcolor{lightgray} \bf in} & {\cellcolor{lightgray} \bf in} \\ \hline
[160]\^ & a & {\cellcolor{lightgray} \bf a} & {\cellcolor{lightgray} \bf a} & {\cellcolor{lightgray} \bf a} \\ \hline
349.[1]= & hurry &  & I & {\cellcolor{lightgray} \bf hurry} \\ \hline
572.[5]- {\tiny +ing} & resuming &  & said & requiring \\ \hline
251.[6]= & every & {\cellcolor{lightgray} \bf every} & {\cellcolor{lightgray} \bf every} & {\cellcolor{lightgray} \bf every} \\ \hline
[852]\^ & point &  & people & precaution \\ \hline
\end{tabular}
\end{center}
\caption{Decipherment of a portion of our evaluation set, compared to the gold standard.}
\label{solution}
\end{figure*}

\section{Conclusion and Future Work}

In this work, we show that it is possible to decipher a book-based cipher, using a known-plaintext attack and a neural English language model. We apply our method to letters written to and from US General James Wilkinson, and we recover 75.1\% of the word tokens correctly.  

We believe word-based neural language models are a  powerful tool for decrypting classical codes and ciphers.  Because they have much lower perplexities than widely-used n-gram models, they can distinguish between candidate plaintexts that resemble English at a distance, versus candidate plaintexts that are grammatical, sensible, and relevant to the historical context.

\section*{Acknowledgments}

We would like to thank Johnny Fountain and Kevin Chatupornpitak of Karga7, and the staff who transcribed data from the Library of Congress, who provided scans of the original documents.  We would also like to thank the anonymous reviewers for many helpful suggestions.

\bibliography{acl2020}
\bibliographystyle{acl_natbib.bst}

\end{document}